\documentclass[runningheads]{llncs}
\usepackage{graphicx}
\usepackage{tikz}
\usepackage{comment}
\usepackage{amsmath,amssymb} % define this before the line numbering.
\usepackage{color}

\usepackage[accsupp]{axessibility}  % Improves PDF readability for those with disabilities.

\usepackage{booktabs}
\usepackage{multirow}
\usepackage{makecell}
\usepackage{dsfont}
\DeclareMathOperator*{\argmax}{arg\,max}
\usepackage{verbatim}
\usepackage{hyperref}
\usepackage{MnSymbol}
\begin{document}
% \linenumbers
\pagestyle{headings}
\mainmatter
\def\ECCVSubNumber{5913}  % Insert your submission number here

\title{Open Vocabulary Object Detection with Pseudo Bounding-Box Labels} % Replace with your title

\begin{comment}
\titlerunning{ECCV-22 submission ID \ECCVSubNumber} 
\authorrunning{ECCV-22 submission ID \ECCVSubNumber} 
\author{Anonymous ECCV submission}
\institute{Paper ID \ECCVSubNumber}
\end{comment}
%******************

%\begin{comment}
\titlerunning{PB-OVD}
\author{Mingfei Gao\thanks{Mingfei and Chen contributed equally.}\ , Chen Xing$^\thinstar$, Juan Carlos Niebles\index{Niebles, Juan Carlos}, Junnan Li, \\ Ran Xu, Wenhao Liu, Caiming Xiong
  }
\authorrunning{Gao et al.}
\institute{Salesforce Research, Palo Alto, USA \\
\email{\{mingfei.gao,cxing,jniebles,junnan.li\}@salesforce.com}
\email{\{ran.xu,wenhao.liu,cxiong\}@salesforce.com}
}
%\end{comment}
%******************
\maketitle

\begin{abstract}
Despite great progress in object detection, most existing methods work only on a limited set of object categories, due to the tremendous human effort needed for bounding-box annotations of training data. To alleviate the problem, recent open vocabulary and zero-shot detection methods attempt to detect novel object categories beyond those seen during training. They achieve this goal by training on a pre-defined base categories to induce generalization to novel objects.
However, their potential is still constrained by the small set of base categories available for training.
To enlarge the set of base classes, we propose a method to automatically generate pseudo bounding-box annotations of diverse objects from large-scale image-caption pairs. 
Our method leverages the localization ability of pre-trained vision-language models to generate pseudo bounding-box labels and then directly uses them for training object detectors. Experimental results show that our method outperforms the state-of-the-art open vocabulary detector by 8\% AP on COCO novel categories, by 6.3\% AP on PASCAL VOC, by 2.3\% AP on Objects365 and by 2.8\% AP on LVIS. \href{ https://github.com/salesforce/PB-OVD}{\emph{Code is available here}}. 

\end{abstract}

\section{Introduction}

\label{sec:intro}
\begin{figure}[t]
\centering
   \includegraphics[width=1\linewidth]{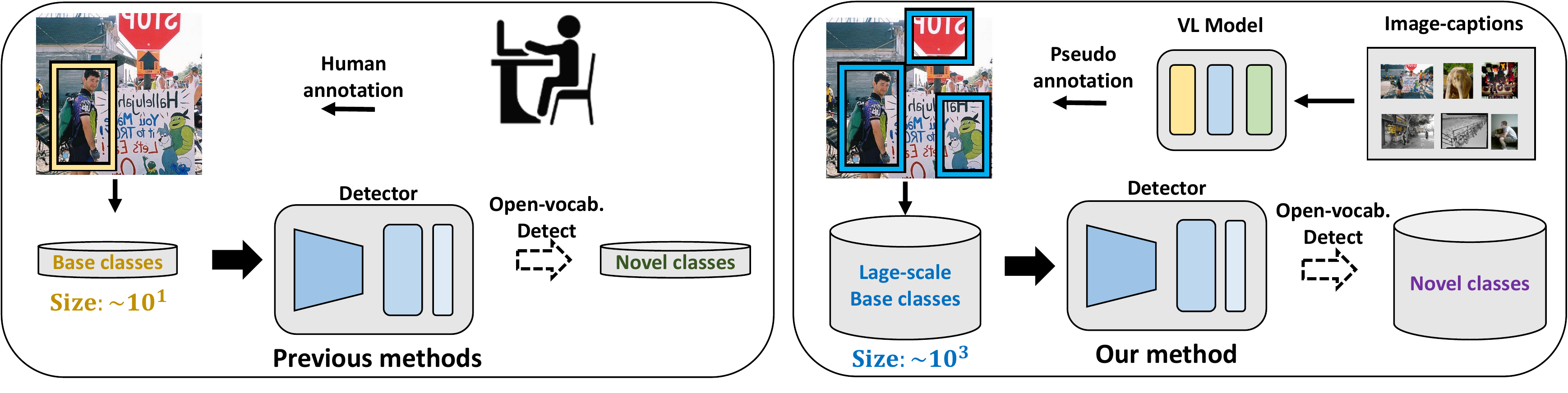}

\caption{Previous methods (left) rely on human-provided box-level annotations of pre-defined base classes during training and attempt to generalize to objects of novel classes during inference. Our method (right) generates pseudo bounding-box annotations from large-scale image-caption pairs by leveraging the localization ability of pre-trained vision-language (VL) models. Then, we utilize them to improve our open vocabulary object detector. Compared to the human annotations of a fixed/small set of base classes, our pseudo bounding-box label generator easily scales to a large set of diverse objects from the large-scale image-caption dataset, thus is able to achieve better detection performance on novel objects compared to previous methods}

\label{fig:fig1} 
\end{figure}

Object detection~\cite{papageorgiou1998general,papageorgiou2000trainable,szegedy2013deep,he2017mask} is a core task in computer vision that has considerably advanced with the adoption of deep learning and continues to attract significant research effort~\cite{sun2021sparse,yin2021center,xie2021detco}. 
Current deep object detection methods achieve astonishing performance
when learning a pre-defined set of object categories that have been annotated in a large number of training images (PASCAL VOC~\cite{everingham2015pascal}, COCO~\cite{lin2014microsoft}).
Unfortunately, their success is still limited to detecting a small number of object categories (e.g., $80$ categories in COCO). 
One reason is that most detection methods rely on supervision in the form of instance-level bounding-box annotations, hence requiring very expensive human labeling efforts to build training datasets. Furthermore, when we need to detect objects from a new category, one has to further annotate a large number of bounding-boxes in images for this new object category. 

Two families of recent work have attempted to reduce the need of annotating new object categories: zero-shot object detection and open vocabulary object detection.
In zero-shot detection methods~\cite{bansal2018zero,rahman2020improved}, object detection models are trained on \textit{base} object categories with human-provided bounding box annotations to promote their generalization ability on \textit{novel} object categories, by exploiting correlations between base and novel categories. These methods can alleviate the need for large amounts of human labeled data to some extent. 
Building on top of such methods, open vocabulary object detection~\cite{zareian2021open}
aims to improve the detection performance of novel objects with the help of image captions.
However, the potential of existing zero-shot and open vocabulary methods is constrained by the small size of the base category set at training, due to the high cost of acquiring large-scale bounding-box annotations of diverse objects. As a result, it is still challenging for them to generalize well to diverse objects of novel categories in practice.

A potential avenue for improvement is to enable open vocabulary detection models to utilize a larger set of base classes of diverse objects by reducing the requirement of manual annotations. In this paper we ask:
can we automatically generate bounding-box annotations for objects at scale using existing resources? Can we use these generated annotations to improve open vocabulary detection? The most recent progress on vision-language pre-training gives us some hope. Vision-language models~\cite{li2019visualbert,tan2019lxmert,radford2021learning,jia2021scaling,ALBEF} are pre-trained with large scale image-caption pairs from the web. They show amazing zero-shot performance on image classification, as well as promising results on tasks related to text-visual region alignment, such as referring expressions, which implies strong localization ability. 

Motivated by these observations, we improve open vocabulary object detection using pseudo bounding-box annotations generated from large-scale image-caption pairs, by taking advantage of the localization ability of pre-trained vision-language models. As shown in Fig. \ref{fig:fig1}, we design a pseudo bounding-box label generation strategy to automatically obtain pseudo box annotations of a diverse set of objects from existing image-caption datasets. 
Specifically, given a pre-trained vision-language model and an image-caption pair, we compute an activation map (Grad-CAM~\cite{selvaraju2017grad}) in the image that corresponds to an object of interest mentioned in the caption. We then convert the activation map into a pseudo bounding-box label for the corresponding object category. Our open vocabulary detector is then directly trained with supervision of these pseudo labels. Our detector can also be fine-tuned with human-provided bounding boxes if they are available.
Since our method for generating pseudo bounding-box labels is fully automated with no manual intervention, the size and diversity of the training data, including the number of training object categories, can be largely increased. This enables our approach to outperform existing zero-shot/open vocabulary detection methods that are trained with a limited set of base categories.

We evaluate the effectiveness of our method by comparing with the state-of-the-art (SOTA) zero-shot and open vocabulary object detectors on four widely used datasets: COCO, PASCAL VOC, Objects365 and LVIS. Experimental results show that our method outperforms the best open vocabulary detection method by 8\% mAP on novel objects on COCO, when both of the methods are fine-tuned with COCO base categories. Moreover, we surprisingly find that even when not fine-tuned with COCO base categories, our method can still outperform 
fine-tuned SOTA baseline by 3\% mAP. We also evaluate the generalization performance of our method to other datasets. Experimental results show that under this setting, our method outperforms existing approaches by 6.3\%, 2.3\% and 2.8\% on PASCAL VOC, Objects365 and LVIS, respectively.

Our contributions are summarized as follows: (1) We propose an open vocabulary object detection method that can train detectors with pseudo bounding-box annotations generated from large-scale image-caption pairs. To the best of our knowledge, this is the first work which enables open vocabulary object detection using pseudo labels during training. (2) We introduce a pseudo label generation strategy using the existing pre-trained vision-language models. (3) With the help of pseudo labels, our method largely outperforms the SOTA methods. Moreover, when trained with only pseudo labels, our method achieves higher performance than the SOTA that rely on training with manual bounding-box annotations.

\section{Related Work}

\noindent\textbf{Object detection} aims at localizing objects in images. Traditional detection methods are supervised using human-provided bounding box annotations.
Two-stage detection methods~\cite{girshick2015fast,ren2015faster,he2017mask} are one of the most popular frameworks. 
These methods generate object proposals in the first stage and classify these proposals to different categories in the second stage. 
Weakly supervised object detectors seek to relieve such heavy human annotation burden by using image-level labels such as image-level object categories~\cite{bilen2016weakly,tang2018pcl}, captions~\cite{ye2019cap2det} and object counts~\cite{gao2018c} for training. Although these approaches show promising performance, they only support objects in a fixed set of categories. Whenever one needs to detect objects from a new category, they have to collect and manually annotate instances from the new category and retrain the detector.

\noindent\textbf{Open vocabulary and zero-shot object detection} target at training an object detector with annotations on base object classes to generalize to novel object classes during inference. Most zero-shot detection methods achieve this level of generalization by aligning the visual and the text representation spaces for objects from base classes during training~\cite{bansal2018zero,rahman2020improved,zhu2020don}, and inferring novel objects during inference by exploiting correlation between base and novel objects. 
Recent methods encourage the visual-semantic alignment for novel objects by different strategies such as synthesizing visual representations of novel classes~\cite{zhu2019zero,zhu2020don} or utilizing existing object names semantically similar to their names~\cite{rahman2020improved}. 
Joseph et al. introduce OREO~\cite{joseph2021towards} to incrementally learn unknown objects based on contrastive clustering and energy based unknown identification.
To further improve the zero-shot performance on novel object categories,  Zareian et al. \cite{zareian2021open} propose open vocabulary object detection that transfers knowledge from a pre-trained vision-language model by initializing their detector with parameters of the image encoder of the vision-language model. This strategy improves the state-of-the-art by a large margin. 
Gu et al. \cite{gu2021zero} propose ViLD which achieves good zero-shot performance by distilling knowledge from a large-scale vision-language model (CLIP~\cite{radford2021learning}).
However, all these methods are trained with a small set of base categories that have bounding-box labels since acquiring bounding-box annotations of diverse objects in large-scale training data is expensive. In practice, if a novel category at inference is very different from base categories, it is still challenging for these methods to generalize well to such novel objects.
In contrast, our method generates pseudo box labels for diverse objects from large-scale image-caption pairs and use them to train the detector. When human-provided box annotations are available, our framework has the flexibility to utilize them.

\noindent\textbf{Vision-language pre-training models} are trained with large-scale image-caption pairs. They have been successful not only in image-language tasks such as image retrieval, VQA and referring expression~\cite{li2019visualbert,tan2019lxmert,jia2021scaling,ALBEF}, but also in pure image tasks such as zero-shot image classification~\cite{radford2021learning}.
Recent methods typically utilize a multi-modal module to encourage the interaction between the vision and language modalities~\cite{li2019visualbert,tan2019lxmert,ALBEF}, which may implicitly encode the word-to-region localization information inside the model. We take advantage of their localization ability and design a strategy to obtain pseudo bounding-box labels of a large and diverse set of objects from the large-scale image-caption datasets.
With this strategy, we largely improve open vocabulary object detection.

\section{Our Approach}
Our framework contains two components: a pseudo bounding-box label generator and an open vocabulary object detector. Our pseudo label generator automatically generates bounding-box labels for a diverse set of objects by leveraging a pre-trained vision-language model. We then train our detector directly with the generated pseudo labels. For fair comparison with existing open vocabulary detection methods, when base object categories with human annotated bounding-boxes are available, we can also fine-tune our trained detector with such data.

\begin{figure*}[th]
\centering
\includegraphics[width=1.0\linewidth]{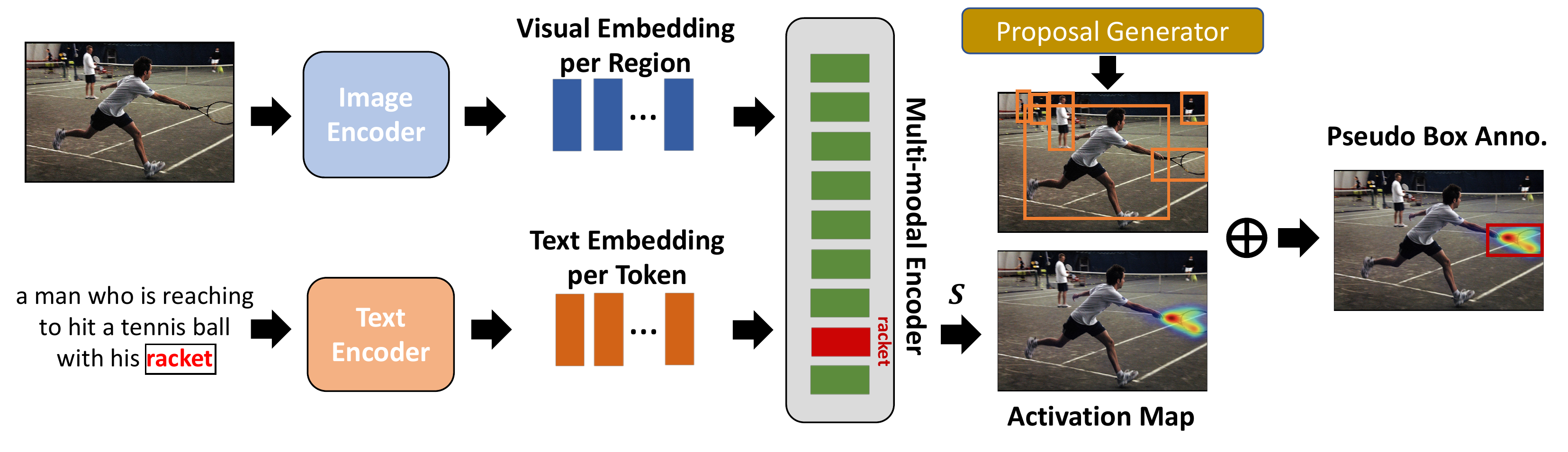}

\caption{Illustration of our pseudo box label generation process. The input to the system is an image-caption pair. We use image and text encoders to extract the visual and text embeddings of the image and its corresponding caption. We then obtain multi-modal features by image-text interaction via cross-attention. We maintain objects of interest in our pre-defined object vocabulary. For each object of interest embedded in the caption (for example, \emph{racket} in this figure), we use Grad-CAM to visualize its activation map in the image. This map indicates the contribution of the image regions to the final representation of the object word. Finally, we determine the pseudo bounding-box label of the object by selecting the object proposal that has the largest overlap with the activation}

\label{fig:fig2} 
\end{figure*}

\subsection{Generating Pseudo Box Labels}
Fig. \ref{fig:fig2} illustrates the overall procedure of our pseudo label generation. 
Our goal is to generate pseudo bounding-box annotations for objects of interest in an image, by leveraging the implicit alignment between regions in the image and words in its corresponding caption in a pre-trained vision-language model.
Before diving into our method, we first briefly introduce the general structure of the recent vision-language models.

An image $\mathbf{I}$ and its corresponding caption, $\mathbf{X} = \{x_1, x_2, ...,x_{N_{T}}\}$, are the inputs to the model, where $N_T$ is the number of words in the caption (including [CLS] and [SEP]). 
An image encoder is used to extract image features $\mathbf{V} \in \mathbb{R}^{N_V \times d}$ and a text encoder is utilized to get text representations $\mathbf{T} \in \mathbb{R}^{N_T \times d}$. $N_V$ is the number of region representations of the image. Moreover, a multi-modal encoder with $L$ consecutive cross-attention layers is often employed to fuse the information from both image and text encoders. 
In the $l$-th cross-attention layer, the interaction of an object of interest $x_t$ in the caption with the image regions is shown in Equation \ref{eq:cross_att} and \ref{eq:cross_att_output}, where $\mathbf{A}_t^{l}$ denotes the corresponding visual attention scores at the $l$-th cross-attention layer. $\mathbf{h}_t^{l-1}$ indicates the hidden representations obtained from the previous $(l-1)$-th cross-attention layer and $\mathbf{h}_t^0$ is the representation of $x_t$ from the text encoder. 

\begin{align}
    \mathbf{A}_t^{l} & =  \text{Softmax}(\frac{\mathbf{h}_t^{l-1}\mathbf{V}^T}{\sqrt{d}}), \label{eq:cross_att} \\
    \mathbf{h}_t^l & =  \mathbf{A}_t^{l}\cdot\mathbf{V}. \label{eq:cross_att_output}
\end{align}

From these equations, a cross-attention layer measures the relevance of the visual region representations with respect to a token in the input caption, and calculates the weighted average of all visual region representations accordingly.  As a result, the visual attention scores $\mathbf{A}_t^{l}$ can directly reflect how important different visual regions are to token $x_t$. Therefore, we visualize the activation maps based on the attention scores to locate an object in an image given its name in the caption.

We use Grad-CAM~\cite{selvaraju2017grad} as the visualization method and follow its original setting to take the final output $s$ from the multi-modal encoder, and calculate its gradient with respect to the attention scores. $s$ is a scalar that represents the similarity between the image and its caption.
Specifically,  the final activation map $\mathbf{\Phi}_t$ of the image given an object name $x_t$ is calculated as
\begin{equation}
    \mathbf{\Phi}_t = \mathbf{A}_t^{l} \cdot \max(\frac{\partial s}{\partial \mathbf{A}_t^{l}}, 0).
    \label{eq:grad}
\end{equation}
If there are multiple attention heads in a cross-attention layer, we average the activation map $\mathbf{\Phi}_t$ from all heads as the final activation map.

After we get an activation map of an object of interest in the caption using this strategy, we draw a bounding box covering the activated region as the pseudo label of the category. We adopt existing proposal generators, e.g.,~\cite{uijlings2013selective,he2017mask} to generate proposal candidates $\mathbf{B}=\{b_1, b_2,...,b_K\}$ and select the one overlapping the most with $\mathbf{\Phi}_t$:

\begin{equation}
\hat{b} = \argmax_{i} \frac{\sum_{b_{i}}\mathbf{\Phi}_t{(b_i)}}{\sqrt{|b_i|}},
\label{eq:box_pick}
\end{equation}
where $\sum_{b_{i}}\mathbf{\Phi}_t{(b_i)}$ indicates summation of the activation map within a box proposal and $|b_i|$ indicates the proposal area. In practice, we maintain a list of objects of interest (referred as object vocabulary) during training and get pseudo bounding-box annotations for all objects in the training vocabulary (see Sec.~\ref{sec:object_vocab} for details).
Fig.~\ref{fig:actmap} shows some examples of the activation maps. As we can see, the activated regions correspond well with the relevant regions. The generated bounding boxes are of good quality. When they are directly used to train an open vocabulary object detector, the object detector significantly outperforms the current SOTA open-vocabulary/zero-shot object detectors.

\subsection{Open Vocabulary Object Detection with Pseudo Labels}

\begin{figure}[t]
\centering
 \includegraphics[width=0.8\linewidth]{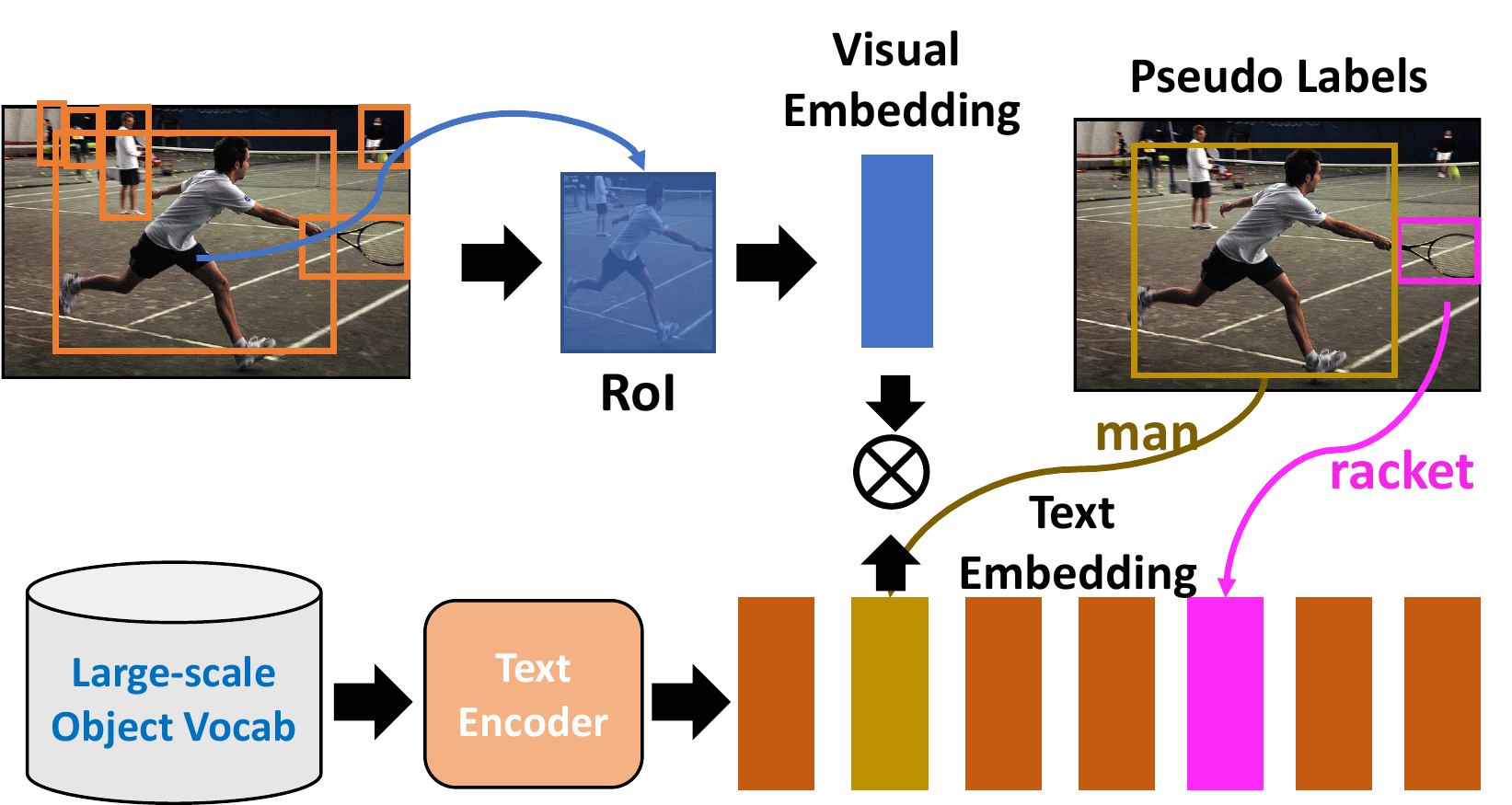}

\caption{Illustration of our detector. An image is processed by a feature extractor followed by a region proposal network. Region-based features are then calculated by applying RoI pooling/RoI align over region proposals and the corresponding visual embeddings are obtained. Similarity of the visual and text embeddings of the same object are encouraged during training}

\label{fig:fig3} 
\end{figure}

After we get pseudo bounding-box labels, we can use them to train an object detector. Since our pseudo-label generation is disentangled from detector training process, our framework can accommodate detectors with any architecture. In this work, we focus on the open vocabulary scenario where a detector aims at detecting arbitrary objects during inference.

A general open vocabulary detection system~\cite{zareian2021open} is shown in Fig.~\ref{fig:fig3}. A feature map is extracted from an input image using a feature extractor based on which object proposals are generated. Then, region-based visual embeddings, $\mathbf{R}=\{\mathbf{r}_1, \mathbf{r}_2,...,\mathbf{r}_{N_r}\}$, are obtained by RoI pooling/RoI align~\cite{he2017mask} followed by a fully connected layer, where $N_r$ denotes the number of regions. In the meanwhile, text embeddings, $\mathbf{C}=\{\mathbf{bg}, \mathbf{c}_1,\mathbf{c}_2,...,\mathbf{c}_{N_{c}}\}$, of object candidates from the object vocabulary are acquired by a pre-trained text encoder, where $N_{c}$ is the training object vocabulary size and $\mathbf{bg}$ indicates ``background'' that matches irrelevant visual regions. The goal of the open vocabulary object detector is to pull close the visual and text embeddings of the same objects and push away those of different objects. The probability of $\mathbf{r}_i$ matches $\mathbf{c}_j$ is calculated as

\begin{equation}
    p(\mathbf{r}_i \ \text{matches} \ \mathbf{c}_j) = \frac{\exp{(\mathbf{r}_i\cdot\mathbf{c}_j})}{\exp(\mathbf{r}_i\cdot \mathbf{bg})+ \sum_k \exp{(\mathbf{r}_i\cdot\mathbf{c}_k})},
\end{equation}
where text embeddings $\mathbf{C}$ is fixed during training. The cross entropy loss is used to encourage the matching of positive pairs and discourage the negative ones.

During inference, given a group of object classes of interest, a region proposal will be matched to the object class if its text embedding has the smallest distance to the visual embedding of the region compared to all object names in the vocabulary. This strategy is similar to other zero-shot/open vocabulary detection methods.
To perform a fair comparison to prior work, we also adopt Mask-RCNN as the base of our open vocabulary detector. We set $\mathbf{bg}=\mathbf{0}$ and include objectness classification, objectness box regression and class-agnostic box regression losses following~\cite{zareian2021open}.

\section{Experiments}
\subsection{Datasets and Object Vocabulary for Training}
\label{sec:object_vocab}
\noindent\textbf{Training Datasets}. 
In our method, we generate pseudo bounding-box annotations of diverse objects from a combination of existing image-caption datasets including COCO Caption~\cite{chen2015microsoft}, Visual-Genome~\cite{krishna2017visual}, and SBU Caption~\cite{ordonez2011im2text}. 
Our final dataset for pseudo label generation and detector training contains about one million images.

\noindent\textbf{Object Vocabulary}. When we generate pseudo labels for object categories from the aforementioned dataset, our default object vocabulary is constructed by the union of all the object names in COCO, PASCAL VOC, Objects365 and LVIS, resulting in 1,582 categories.
We would also like to note that since our method doesn't require extra human annotation efforts, our training object vocabulary can be easily augmented. 

\subsection{Evaluation Benchmarks}
\label{sec:evaluation}
\noindent\textbf{Baselines}. We compare with recent zero-shot and open vocabulary methods \cite{bansal2018zero,zhu2020don,rahman2020improved,zareian2021open}. Among the baselines, \emph{Zareian et al.}~\cite{zareian2021open} is the SOTA method, thus, is treated as our major baseline.

\noindent\textbf{Generalized Setting in COCO}. Most existing methods are evaluated under this setting proposed in \cite{bansal2018zero}. COCO detection training set is split to base set containing 48 base/seen classes and target set including 17 novel/unseen classes. Base classes are used for training. During inference, models predict object categories from the union of base and novel classes. The performance of models is evaluated using the mean average precision over the novel classes. 

Our method can be trained using the large-scale dataset with the generated pseudo labels. To perform a fair comparison with baselines, we fine-tune our detector using COCO base categories following their setup. Moreover, we also report our method's performance without fine-tuning on COCO base categories.

\noindent\textbf{Generalization Ability to Other Datasets}. We are interested in measuring the generalization ability of a model to other datasets that the model is not trained on. Therefore, we evaluate our method and the strongest baseline (both are fine-tuned using COCO base classes) on PASCAL VOC~\cite{everingham2007pascal} test set, Objects365 v2~\cite{shao2019objects365} validation set and LVIS~\cite{gupta2019lvis} validation set~\footnote{We use LVIS v0.5, since the validation set of LVIS v1.0 contains images from \emph{COCO train 2017} which our method may finetune on in some experiments.}.
%We directly use their object names as the candidates in the inference vocabulary. 
PASCAL VOC is a widely used dataset by traditional object detection methods which contains 20 object categories. Objects365 and LVIS are datasets include 365 and 1,203 object categories, respectively, which makes them very challenging. When evaluating on each of these datasets (PASCAL VOC, Objects365 and LVIS), visual regions will be matched to one of the object categories (including background) of each dataset during inference.

\noindent\textbf{Evaluation Metric}. Following prior work~\cite{zareian2021open}, we use the standard metric in object detection tasks, i.e., mean average precision over classes, and set the IoU threshold to 0.5. 

\begin{table*}[tb]
    \centering
        \caption{Performance on COCO dataset. Our method outperforms all the previous approaches when all models are fine-tuned on COCO base categories. When our method is not fine-tuned, it still outperforms other fine-tuned baselines}
    \label{tab:coco_zeroshot}
    \begin{tabular}{l|c|c|>{\color{gray}\arraybackslash}c|>{\color{gray}\arraybackslash}c}
    \toprule
         &Fine-tuned with Box Anno. &\multicolumn{3}{c}{Generalized Setting} \\
         Method&  \makecell[c]{ on COCO \textbf{Base} Categories } & Novel AP & Base AP & Overall AP  \\
         \hline
         Bansal et al.~\cite{bansal2018zero}& Yes & 0.3 & 29.2 &24.9 \\
         Zhu et al.~\cite{zhu2020don}& Yes & 3.4 & 13.8 & 13.0 \\
         Rahman et al.~\cite{rahman2020improved}& Yes & 4.1 & 35.9 & 27.9 \\
         \hline
         Zareian et al.~\cite{zareian2021open}& Yes & 22.8 & 46.0 & 39.9 \\
         \hline
        Our method & Yes& \textbf{30.8}& 46.1& 42.1\\ \hline\midrule
         Our method& No & \textbf{25.8}& --&--\\
    \bottomrule
    \end{tabular}
\end{table*}
\begin{table}[tb]
    \centering
    \caption{Generalization performances to other datasets. Our method has better generalization performance to other datasets compared to \emph{Zareian et al.}}
    \label{tab:generalization_to_other_datasets}
    \begin{tabular}{l|c|c|c|c}\toprule
         Method & Fine-tuned on COCO  &PASCAL VOC &  Objects365 & LVIS \\
         &\makecell[c]{ \textbf{Base} Categories } &&& \\
         \hline
        Zareian~\cite{zareian2021open} &  Yes &52.9 & 4.6&5.2 \\
         \hline
         Our method & Yes & \textbf{59.2}& \textbf{6.9}&\textbf{8.0}\\\hline\midrule
         Our method & No & 44.4& \textbf{5.1}& \textbf{6.5}\\
    \bottomrule
    \end{tabular}

\end{table}

\subsection{Implementation Details}
\label{sec:implementation_details}
In our main experiment, we use the ALBEF model pre-trained with 14M data~\footnote{https://github.com/salesforce/ALBEF (BSD-3-Clause License)} as our vision-language model for pseudo label generation. We follow the default setting of ALBEF, unless otherwise noted. The cross-attention layer used for Grad-CAM visualization is set to $l=8$ in Eq.~\ref{eq:grad}. We conduct our main experiments using ALBEF because of its good performance in object grounding when image captions are present. Note that other pre-trained vision-language models can also fit our framework without major modifications or adding additional constraints on detector training. We conduct an ablation study to show our method's performance when another pre-trained vision-language model (LXMERT~\cite{tan2019lxmert}) is employed in Sec.~\ref{sec:ablation}.
As a default option, we use a off-the-shelf Mask-RCNN with ResNet-50 trained on COCO 2017 train set as our proposal generator. To ensure there is no labels of novel categories leaking to our model, we have excluded the novel categories when training the proposal generator. 
We also show our results with an unsupervised proposal generator, selective search~\cite{uijlings2013selective}, in Sec.~\ref{sec:ablation}.

We use Mask-RCNN with ResNet-50 as the base of our open vocabulary detector and keep following the default settings here~\footnote{https://github.com/alirezazareian/ovr-cnn (MIT License)}. We utilize the pre-trained CLIP (ViT-B/32) text encoder to extract text embeddings of objects in our vocabulary and use the text prompts provided in~\cite{gu2021zero} to ensemble the text representation.
We use a batch size of 64 with learning rate of 0.02 when training the open vocabulary detector using our large-scale dataset with pseudo bounding-box labels, and a batch size of 8 with base learning rate of 0.0005 when optionally fine-tuning on COCO base classes. Models are optimized using SGD. The weight decay is set to 0.0001. The maximum iteration number is 150,000 and the learning rate is updated by a decreasing factor of 0.1 at 60,000 and 120,000 iterations.

\begin{table}[tb]
    \centering
    \caption{Effect of proposal quality. All models are fine-tuned on COCO base classes. Better proposals lead to better detection performance}
    \label{tab:proposal_ablation}
    \begin{tabular}{l|c|c|c|c|c}\toprule
          Method & \makecell[c]{Proposal \\ Generator} &  \makecell[c]{Generalized \\ COCO Novel} &PASCAL VOC & Objects365 & LVIS\\
         \hline
         Zareian et al.\cite{zareian2021open}& -- & 22.8&52.9 & 4.6 & 5.2 \\ 
         \hline
          \multirow{2}{*}{Our method} & {Selective Search}& \textbf{28.5}& \textbf{53.0}& \textbf{5.5}& \textbf{5.9}\\
          &Mask-RCNN & \textbf{30.8}&\textbf{59.2} & \textbf{6.9}& \textbf{8.0}\\
         \bottomrule
    \end{tabular}

\end{table}

\begin{table}[tb]
    \centering
        \caption{Effect of different vision-language models. All models are fine-tuned on COCO base classes. Better VL model leads to better detection performance}
    \label{tab:vl_ablation}
    \begin{tabular}{l|c|c|c|c|c}\toprule
          Method & VL Model &  \makecell[c]{Generalized \\ COCO Novel} &PASCAL VOC & Objects365 & LVIS\\
         \hline
         Zareian et al.\cite{zareian2021open}& -- & 22.8&52.9 & 4.6 & 5.2 \\ 
         \hline
          \multirow{2}{*}{Our method} & LXMERT~\cite{tan2019lxmert}&  \textbf{27.0} & \textbf{56.5} & \textbf{5.5} &\textbf{6.4}\\
          &ALBEF\cite{ALBEF} & \textbf{30.8}&\textbf{59.2} & \textbf{6.9}& \textbf{8.0}\\
         \bottomrule
    \end{tabular}

\end{table}

\subsection{Experimental Results}
\label{sec:result}
As shown in Table~\ref{tab:coco_zeroshot}, when fine-tuned using COCO base categories same as our baselines, our method outperforms our strongest baseline (\emph{Zareian et al.}) largely by 8\%.
When not fine-tuned using COCO base categories and only trained with generated pseudo labels, our method achieves $25.8\%$ AP on the novel categories, which still significantly outperforms the SOTA method (\emph{Zareian et al.}) by 3\%. 

Generalization ability to a wide range of datasets is also important for an open vocabulary object detector, since it makes a detector directly usable as an out-of-the-box method in the wild. Table~\ref{tab:generalization_to_other_datasets} shows the generalization performance of detectors to different datasets, where both our method and our baseline are not trained using these datasets. Since objects365 and LVIS have a large set of diverse object categories, evaluation results on these datasets would be more representative to demonstrate the generalization ability. Results show that our method achieves better performance than \emph{Zareian et al.} on all three datasets when both of the methods are fine-tuned with COCO base categories. Our method improves the results of our baseline by 2.3\% in Objects365 and 2.8\% on LVIS. Besides, our fine-tuned method beats the SOTA largely by 6.3\% on PASCAL VOC. 
When not fine-tuned with COCO base categories, the performance of our method still outperforms \emph{Zareian et al.} (fine-tuned with COCO base categories) on Objects365 and LVIS. 
When not fine-tuned, our method's underperforms our fine-tuned baseline on PASCAL VOC. It is very likely because of that there is a large semantic overlap between the COCO base categories and PASCAL VOC object categories. Therefore, fine-tuning on COCO base set helps the model's transfer ability to PASCAL VOC.

\subsection{Ablation Study}
\label{sec:ablation}
\noindent\textbf{How does the quality of bounding-box proposals affect performance?} 
Our pseudo label generator combines object proposals and the activation map to select boxes. Generally, the better the proposals are, the more accurate our pseudo bounding-box annotations would be.
The default proposal generator in our main experiments is a Mask-RCNN trained with COCO detection category excluding the novel categories. 
To analyze the effect of proposal quality, we also run experiments with an unsupervised proposal generator, Selective Search \cite{uijlings2013selective} and summarize results in Table \ref{tab:proposal_ablation}.
The results show that our method with  Selective Search outperforms \emph{Zareian et al.} with a clear margin.
This demonstrates the effectiveness of our method even with an unsupervised proposal generator. 

\begin{table}[tb]
    \centering
        \caption{Performance of our method when using vocabularies of different sizes for pseudo label generation. $\mathbb{V}^-$ and $\mathbb{V}$ contain 65 and 1.5k+ categories, respectively}
    \label{tab:vocab_ablation}
    \begin{tabular}{l|c|c|>{\color{gray}\arraybackslash}c|>{\color{gray}\arraybackslash}c}\toprule
    & &\multicolumn{3}{c}{Generalized Setting} \\
         Methods & Vocabulary & \makecell[c]{Novel AP\\ (17)} & \makecell[c]{Base AP\\ (48)} & \makecell[c]{Overall AP \\ (65)}\\
         \hline
         \hline
         Zareian et al.\cite{zareian2021open}& --& 22.8& 46.0& 39.9 \\
         \hline
          \multirow{2}{*}{Our method}&$\mathbb{V}^-$(65) & \textbf{29.7}& 44.3& 40.4\\
           &$\mathbb{V}$ (1.5k+)& \textbf{30.8}& 46.1& 42.1 \\
         \bottomrule
    \end{tabular}
\end{table}

\begin{table}[tb]
    \centering
        \caption{Performance of our method when trained with pseudo labels generated from different amounts of data. All models are fine-tuned using COCO base categories. Our performance improves when trained with pseudo labels of more data}
    \label{tab:pretrain_data_ablation}
    \begin{tabular}{l|c|c|>{\color{gray}\arraybackslash}c|>{\color{gray}\arraybackslash}c}\toprule
        & &\multicolumn{3}{c}{Generalized Setting} \\
         Methods & Data of Pseudo Label Generation & \makecell[c]{ Novel AP\\ (17)} & \makecell[c]{ Base AP\\ (48)} & \makecell[c]{ Overall AP \\ (65)}\\
         \hline
          \makecell[c]{Zareian et al.\cite{zareian2021open}} & {--} & 22.8& 46.0& 39.9\\
         \hline
         \multirow{2}{*}{Our method} & { COCO Cap} & \textbf{29.1}& 44.4& 40.4\\
         & { COCO Cap, VG, SBU} & \textbf{30.8}& 46.1& 42.1\\
         \bottomrule
    \end{tabular}
\end{table}

\begin{table}[tb]
    \centering
        \caption{Performance of our method with different text encoders}
    \label{tab:text_embed_ablation}
    \begin{tabular}{l|c|c|>{\color{gray}\arraybackslash}c|>{\color{gray}\arraybackslash}c}\toprule
        & &\multicolumn{3}{c}{Generalized Setting} \\
         Method&Text Encoder & \makecell[c]{Novel AP} & Base AP & Overall AP \\
         %&PASCAL VOC &Objects365&LVIS\\
         \hline
         Zareian et al.\cite{zareian2021open}& Bert & 22.8 & 46.0 & 39.9 \\
         %&52.9 & 4.6 & 5.2 \\
         \hline
           \multirow{2}{*}{Our method}&Bert & \textbf{28.8} & 45.1& 40.9\\
           %&49.4  & \textbf{5.8}& \textbf{5.3}\\
          &CLIP &\textbf{30.8} & 46.1 & 42.1\\
          %&\textbf{59.2} & \textbf{6.9}& \textbf{8.0}\\
         \bottomrule
    \end{tabular}

\end{table}
\begin{figure}[tb]
    \centering
    \includegraphics[width=1.0\linewidth]{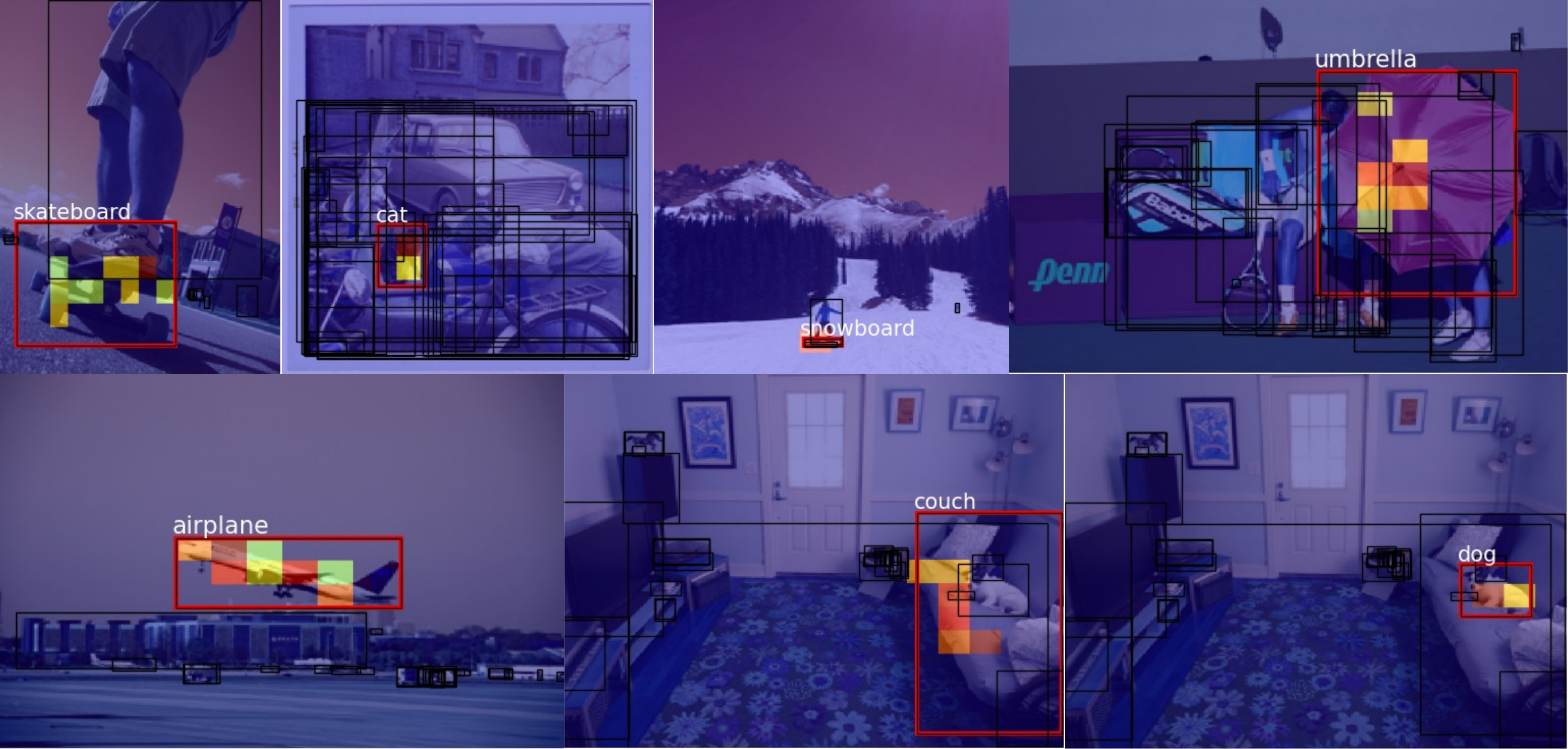}
    \caption{Visualization of some activation maps. Colorful blocks indicate values of Grad-CAM activation map in the corresponding regions. We zero out blocks with values smaller than half of the max
    value in the map so the main focus is highlighted. Black boxes indicate object proposals and read boxes indicate the final selected pseudo bounding-box labels}
    \label{fig:actmap}
\end{figure}
\noindent\textbf{What is the effect of different pre-trained vision-language models?} 
Besides the utilization of ALBEF in our main experiment, we also experiment with LXMERT~\cite{tan2019lxmert}, which is an earlier vision-language model that also fuses information from both the vision and language modalities. Specifically, we generate pseudo labels based on the activation map in the last layer of the (text-to-vision) cross-attention module. Results are shown in Table~\ref{tab:vl_ablation}. It shows that with LXMERT, our method's performance is  slightly worse compared with our method using ALBEF. This may due to the fact that LXMERT employs less image-caption data for training. While compared with \emph{Zareian et al.}, our method with LXMERT still performs significantly better.

\noindent\textbf{What is the effect of our object vocabulary size?}
We utilize an object vocabulary containing over 1.5k object categories ($\mathbb{V}$) by default when generating our pseudo labels. The vocabulary size is much larger than the object  vocabulary of any dataset we evaluated on, such as base and novel object categories in COCO ($\mathbb{V}^-$). 
A natural question is, for performance of COCO novel categories, for example, would it be better if we just use $\mathbb{V}^-$ for pseudo-box generation and detector training? To answer the question, we conduct experiments using these two scales of vocabularies. Experimental results in Table~\ref{tab:vocab_ablation} show that $\mathbb{V}$ leads to better performance than $\mathbb{V}^-$ on COCO novel categories. The results suggest that adding additional object categories outside the COCO categories during pre-training will benefit our model performance. Besides, using the larger vocabulary improves model's generalization ability to datasets that include a large set of object categories, i.e., Objects365 and LVIS. We observe that using $\mathbb{V}$ improves results of $\mathbb{V}^-$ by 2.3\% in Objects365 and by 2.7\% in LVIS.

\noindent\textbf{Does more data for pseudo label generation help?} 
Our pseudo labels are generated from image-caption pairs. Intuitively, the more data is used for pseudo label generation, the larger amount of diverse objects will be utilized for training our open vocabulary detector. As a result, our model performance should be improved. We show the performance of our method using pseudo labels with different amounts of image-caption pairs in Table~\ref{tab:pretrain_data_ablation}.
Results show that our method can benefit from a larger dataset. 
Our performance is improved by $\sim$2\% on the target set when using more data. 
Moreover, our method still outperforms the baseline significantly even when pre-trained with \emph{COCO Cap} only.

\noindent\textbf{What is the effect of different text encoders?}
Besides our default choice of text encoder (CLIP), we implement our method with another encoder, i.e., Bert (base)~\cite{devlin2018bert} which is a widely used language model that is trained using text data only.  Our main baseline \emph{Zareian et al.} uses Bert as their text encoder as well. The comparison results are shown in Table~\ref{tab:text_embed_ablation}. The results suggest that with Bert encoder, our method's performance is  slightly worse  on COCO target set compared with our method using CLIP text encoder. This may due to the fact that CLIP text encoder is trained using image-caption pairs which results in better generalization performance for image-related tasks. While our method with Bert encoder still outperforms \emph{Zareian et al.} by 6\% AP on novel categories. It indicates that the performance improvement of our method doesn't mainly come from a better text encoder, but from training with large-scale pseudo bounding-box labels.

\begin{figure*}[t]
\centering
\includegraphics[width=1.0\linewidth]{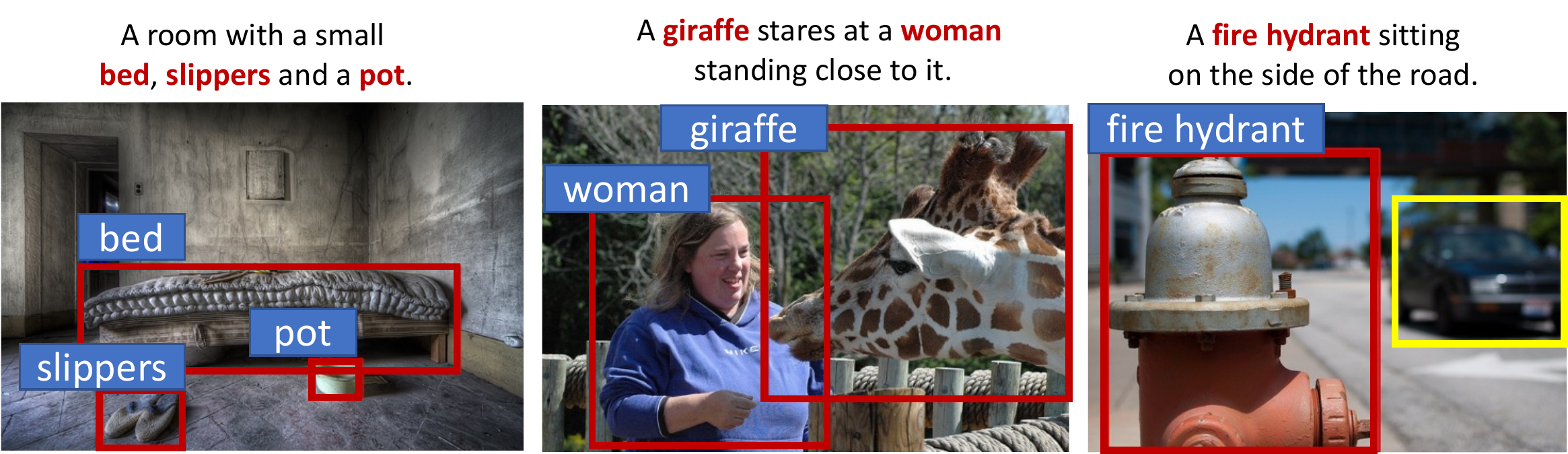}

\caption{Visualization of our generated pseudo bounding-box annotations on COCO. The red boxes indicate successful cases and the yellow one denote failure case. Our pseudo label generator can generate objects (slippers, pot and pie) that are not covered by COCO's category list. 
The generator cannot capture an object if it is not shown in the caption (e.g. the car in the last column)}
\label{fig:plabels_qualitative} 

\end{figure*}

\begin{figure*}[th]
\centering
\includegraphics[width=1.0\linewidth]{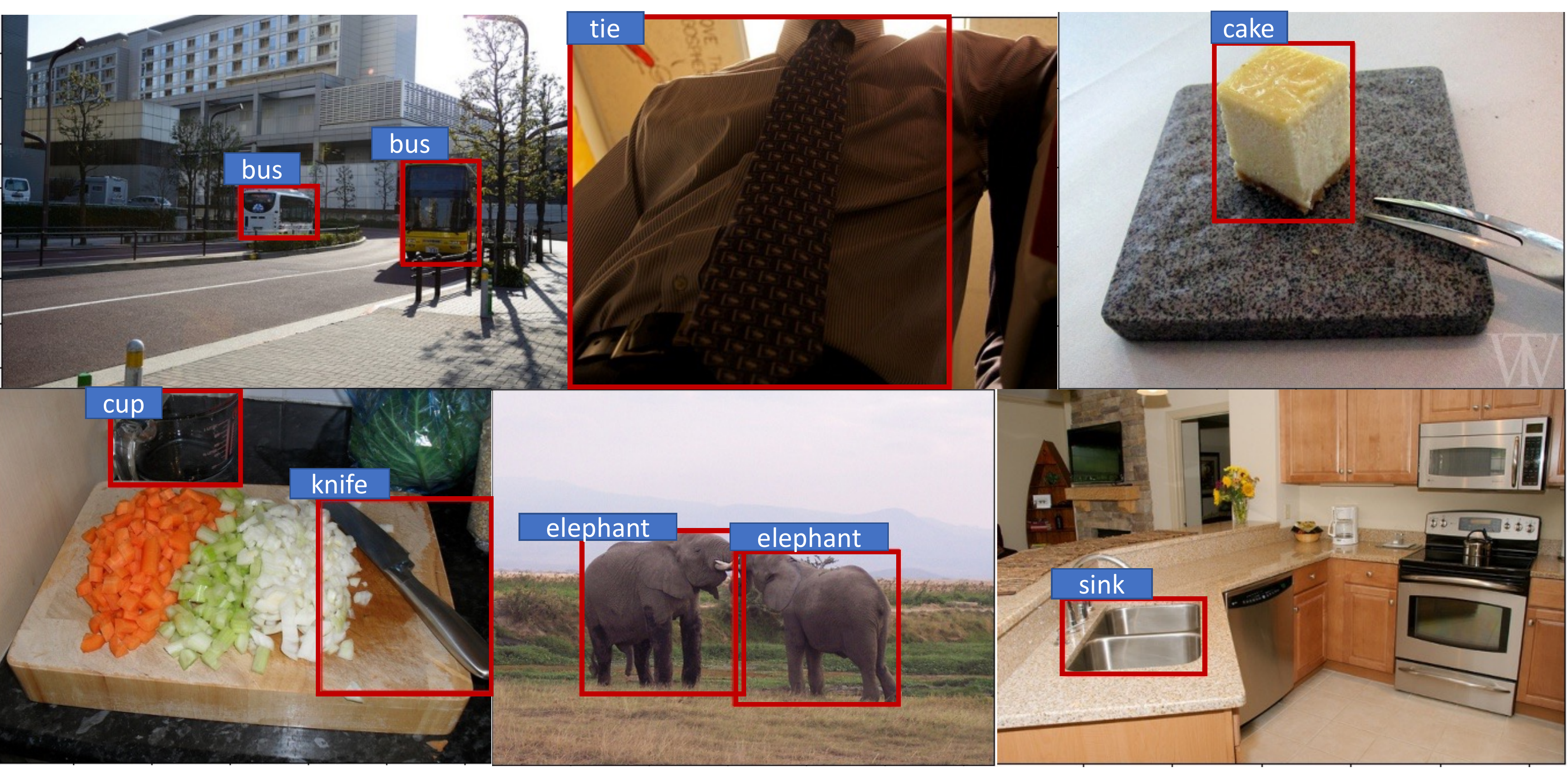}

\caption{Some example results of our open vocabulary detector. The shown categories are from novel categories in COCO}
\label{fig:ovd_qualitative} 

\end{figure*}

\noindent\textbf{Qualitative visualization.}
\label{sec:experiment_plabel_quality}
We visualize examples of our generated pseudo bounding boxes in Fig.~\ref{fig:actmap} and Fig.~\ref{fig:plabels_qualitative}. As we can see, the generated pseudo labels show promising performance (see red boxes) in grounding objects and are able to cover categories, e.g., pot, slippers and pie (Fig.~\ref{fig:plabels_qualitative}), that are not in the original object list of COCO's ground-truth annotations.
We also observe that some background objects will be missed when it is not mentioned in the caption (see the yellow box in the last column of Fig.~\ref{fig:plabels_qualitative}).
Nevertheless, instances of the same object category may show up in other image-caption pairs in the large-scale dataset and our method could generate pseudo box labels for those cases. Therefore, our detector can still recognize this object after training. We also show some examples of our detection results in Fig.~\ref{fig:ovd_qualitative}. They are all from COCO novel categories which are not covered by COCO base annotations. The results demonstrate promising generalization ability of our method to novel objects.

\section{Closing Remarks}
We propose a novel framework that trains an open vocabulary object detector with pseudo bounding-box labels generated from large-scale image-caption pairs. We introduce a pseudo label generator that leverages the localization ability of pre-trained vision-language models to generate pseudo bounding-box annotations for diverse objects embedded in image captions. The generated pseudo labels can be used to improve open vocabulary object detection. Experimental results show that our method outperforms the state-of-the-art zero-shot and open vocabulary object detection methods by a large margin.

\noindent\textbf{Potential Negative Societal Impact}. Our method generates pseudo bounding box labels to alleviate human labeling efforts. Since our pseudo label generator mines annotations of objects from the input captions without human intervention, our pseudo labels might be biased because of the bias embedded in the language descriptions. Manually filtering out the biased object names in the vocabulary could be an effective solution.

\clearpage
\bibliographystyle{splncs04}
\bibliography{egbib}

\appendix

\section{Additional Analysis}
We conduct the following analysis to understand more about the performance of our method and our major baseline, \emph{Zareian et al.}, \cite{zareian2021open}.

\noindent\textbf{The effect of data amount for pre-training vision-language model}. Both \emph{Zareian et al.} and our method leverage a vision-language (VL) model. Intuitively, a VL model can be improved by training with more data that will potentially improve both our method and baseline. We verify this hypothesis as follows.

First, we attempt to improve \emph{Zareian et al.} by training their VL model with more data. We observe that the performance of \cite{zareian2021open} drops when pre-training with more data. This observation aligns well with the findings in their original paper (see Table 2 of \cite{zareian2021open}). Specifically, \cite{zareian2021open} is originally pre-trained with COCO Caption (the baseline we compared with in our main paper).
When pre-trained with COCO Caption, VG and SBU (1M
images in total), their performance drops by 4.2\% on COCO
novel categories. When pre-trained with COCO Caption,
VG, SBU and CC3M (4M images in total), their performance drops further by 10.8\% compared to our baseline. 

Then, we also try to weaken our method by re-training
our VL model (ALBEF) with COCO caption data only.
The comparison is shown in Table A1. The results suggest that our method still outperforms our baseline largely
by 5\% even when we do not leverage one main strength of
our method: being able to effectively employ more image-caption data without much extra cost.

Also, our method can benefit from more image-caption
data, as suggested by our performance in Table~\ref{tab:our_weaken} (27.8)
vs. our performance (30.8) in Table~\ref{tab:coco_zeroshot} in our main paper.

\begin{table}[h]
    \centering
    \caption{Results when our method only utilizes COCO caption to pre-train our VL model}
    \begin{tabular}{l|c|c}\toprule
         Methods & Pre-train Source & \makecell[c]{COCO Novel AP}\\
         \hline
          Zareian et al. & {COCO Caption} 
          & 22.8\\
          \hline
         Our method & {COCO Caption} & 27.8 \\
         \hline
    \end{tabular}
    \label{tab:our_weaken}
\end{table}

\noindent\textbf{Why \emph{Zareian et al.} cannot easily take advantage of more powerful VL models?} We find that \emph{Zareian et al.} cannot easily incorporate more powerful VL models, e.g., ALBEF~\cite{ALBEF} due to its core designs.  First, it requires the same model architecture for the VL model’s visual encoder and the detection backbone, since the main idea of \cite{zareian2021open} is to initialize their open vocabulary detector using the parameters of visual encoder of the VL model as a way of knowledge transfer. Therefore, their ResNet-based detector makes it impossible to utilize ALBEF’s visual encoder because ALBEF’s
visual encoder is transformer-based. It is also challenging to utilize other SOTA VL models due to the different designs and sizes of visual encoders. In contrast, our framework can utilize most of recent and powerful VL models because they are disentangled from the training of our detectors. This flexibility brings large performance improvement.
\end{document}